\documentclass[runningheads]{llncs}
\usepackage{graphicx}
\usepackage{amsmath,amssymb} 
\usepackage{color}
\usepackage[dvipsnames]{xcolor}
\usepackage[width=122mm,left=12mm,paperwidth=146mm,height=193mm,top=12mm,paperheight=217mm]{geometry}
\usepackage[colorlinks,linkcolor=red, filecolor=black, urlcolor=purple, citecolor=RoyalBlue]{hyperref}
\usepackage{cite}
\usepackage{booktabs}
\usepackage{pifont}%
\usepackage{multirow}
\usepackage{gensymb}
\usepackage{floatrow}
\usepackage{pifont}%
\newcommand{\cmark}{\ding{51}}%
\newfloatcommand{capbtabbox}{table}[][\FBwidth]
\newfloatcommand{figurebox}{figure}[\nocapbeside][\dimexpr(\textwidth-\columnsep)/2\relax]
\newfloatcommand{tablebox}{table}[\nocapbeside][\dimexpr(\textwidth-\columnsep)/2\relax]

\begin{document}

	\pagestyle{headings}
	\mainmatter
	\def\ECCV16SubNumber{755}  
	
	\title{2DPASS: 2D Priors Assisted Semantic Segmentation on LiDAR Point Clouds} 
	
	\titlerunning{2D Priors Assisted Semantic Segmentation on LiDAR Point Clouds}
	
	\authorrunning{X. Yan et al.}
	
\author{Xu Yan$^{1\dagger}$, Jiantao Gao$^{2\dagger}$, Chaoda Zheng$^{1\dagger}$,  \\Chao Zheng$^{3}$, Ruimao Zhang$^{1}$, Shuguang Cui$^{1}$,  Zhen Li$^{1}$\thanks{{ Corresponding author: Zhen Li. $^\dagger$ Equal first authorship.}} 
}
\institute{$^{1}$The Future Network of Intelligence Institute, The Chinese University of \\ Hong Kong (Shenzhen), Shenzhen Research Institute of Big Data,\\ 
	$^{2}$Shanghai University, $^{3}$Tencent Map, T Lab\\
}
	
	\def\etal{\textit{et.al.}}
	\def\eg{{\textit{e.g.}}}
	\def\ie{{\textit{i.e.}}}
	
	\maketitle
	\begin{abstract}
		As camera and LiDAR sensors capture complementary information in autonomous driving, great efforts have been made to conduct semantic segmentation through multi-modality data fusion.
		However, fusion-based approaches require paired data, \ie, LiDAR point clouds and camera images with strict point-to-pixel mappings, as the inputs in both training and inference stages.
		It seriously hinders their application in practical scenarios.
		Thus, in this work, we propose the {2D Priors Assisted Semantic Segmentation} ({\textbf{2DPASS}}) method, a general training scheme, to boost the representation learning on point clouds.
		The proposed 2DPASS method fully takes advantage of 2D images with rich appearance during training, and then conduct semantic segmentation without strict paired data constraints.
		In practice, by leveraging an auxiliary modal fusion and multi-scale fusion-to-single knowledge distillation (MSFSKD), 2DPASS acquires richer semantic and structural information from the multi-modal data, which are then distilled to the pure 3D network.
		As a result, our baseline model shows significant improvement with only point cloud inputs once equipped with the 2DPASS. 
		Specifically, it achieves the state-of-the-arts on two large-scale recognized benchmarks (\ie, SemanticKITTI and NuScenes), \ie, ranking the top-1 in both single and multiple scan(s) competitions of SemanticKITTI.
		Code will be made available at \url{https://github.com/yanx27/2DPASS}.
		
		\keywords{Semantic Segmentation, Multi-Modal, Knowledge Distillation, LiDAR Point Clouds}
		
	\end{abstract}

	\section{Introduction}
	
	Semantic segmentation plays a crucial role in large-scale outdoor scene understanding, which has broad applications in autonomous driving and robotics~\cite{hu2019randla,yan2021sparse,SparseConv}.
	In the past few years, the research community has devoted significant effort to understanding natural scenes using either camera images~\cite{chen2017deeplab,chen2017rethinking,song2017semantic,huang2019ccnet} or LiDAR point clouds~\cite{yan2021sparse,xu2020squeezesegv3,zhu2021cylindrical,tang2020searching,zheng2022beyond,zheng2021box} as the input.
	However, these single-modal methods inevitably face challenges in complex environments due to the inherent limitations of the input sensors.
	Concretely, cameras provide dense color information and fine-grained texture, but they are ambiguous in depth sensing and unreliable in low light conditions. In contrast, LiDARs robustly offer accurate and wide-ranging depth information regardless of lighting variances but only capture sparse and textureless data.
	Since cameras and LiDARs complement each other, it is better to perceive the surrounding with both sensors.
	
\begin{figure*}[t]
		\begin{centering}
			\includegraphics[width=0.9\textwidth]{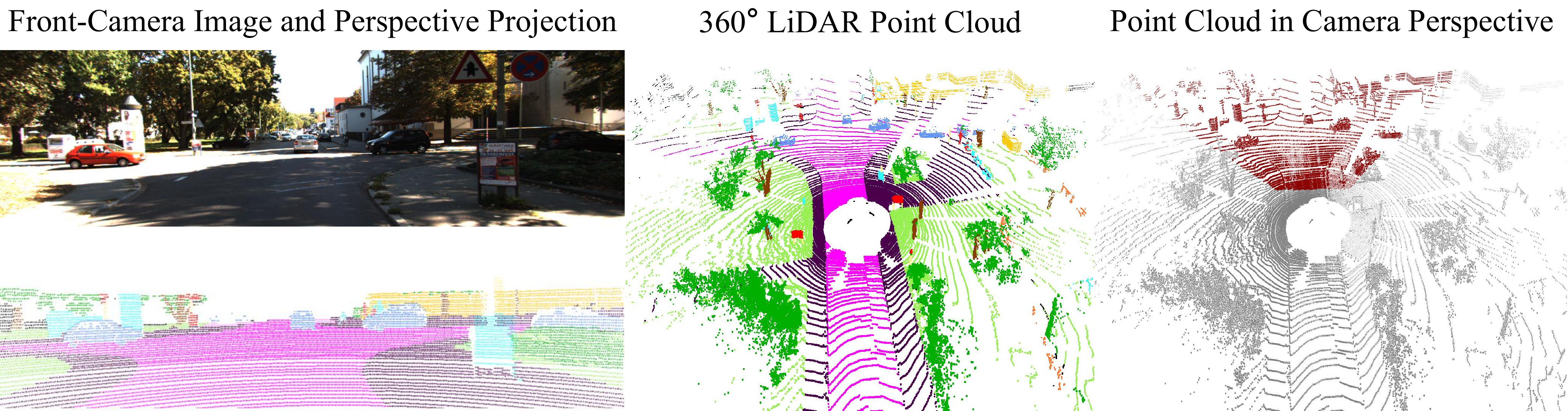}
			\caption{\textbf{Limitation of fusion-based methods.} When the self-driving car only has front-cameras with limited perspective such as SemanticKITTI~\cite{behley2019semantickitti} dataset while the 360-degree LiDAR has a much larger sensing range, fusion-based methods that require strict alignment between camera and LiDAR can only identify a small proportion of the point cloud (see the red region).
			}
			\label{fig:fig1}
		\end{centering}	
	\end{figure*}

	Recently, many commercial cars have been equipped with both cameras and LiDARs. This excites the research community to improve the semantic segmentation by fusing the information from two complementary sensors~\cite{zhuang2021perception,el2019rgb,vora2020pointpainting}.
	These approaches first establish the mapping between 3D points and 2D pixels by projecting the point clouds onto the image planes using the sensor calibrations.
	Based on the point-to-pixel mapping, the models fuse the corresponding image features into the point features, which are further processed to obtain the final semantic scores. 
	Despite the improvements, fusion-based methods have the following unavoidable limitations:
	\textbf{1)} Due to the difference of FOVs (field of views) between cameras and LiDARs, the point-to-pixel mapping cannot be established for points that are out of the image planes. Typically, the FOVs of LiDAR and cameras only overlap in a small portion (see {Fig.~\ref{fig:fig1}}), which significantly limits the application of fusion-based methods.
	\textbf{2)} Fusion-based methods consume more computational resources since they process both images and point clouds (through multitask or cascade manners) at runtime, which introduces a great burden on real-time applications.

	To address the above two issues, we focus on improving semantic segmentation by leveraging both images and point clouds through an effective design in this work.
	Considering the sensors are moving in the scenes, the non-overlap part of the 360-degree LiDAR point clouds corresponding to image in the same time-stamp (see the gray region of the right part in Fig.~\ref{fig:fig1}) can be covered by images from other time-stamp.
	Besides, the dense and structural information of images provides useful regularization for both seen and unseen point cloud regions.
	Based on these observations, we propose a ``model-independent" training scheme, namely {2D Priors Assisted Semantic Segmentation} (\textbf{2DPASS}), to enhance the representation learning of any 3D semantic segmentation networks with minor structure modification.
	In practice, on the one hand, for above-mentioned non-overlap regions, 2DPASS takes pure point clouds as the inputs to train the segmentation model. 
	On the other hand, for subregions with well-aligned point-to-pixel mappings, 2DPASS adopts an auxiliary multi-modal fusion to aggregate image and point features in each scale, and then aligns the 3D predictions with the  fusion predictions.
	Unlike previous cross-modal alignment~\cite{jaritz2020xmuda} apt to contaminate the modal-specific information, we design a multi-scale fusion-to-single knowledge distillation (MSFSKD) strategy to transfer extra knowledge to the 3D model as well as retaining its modal-specific ability.
	Compared with fusion-based methods, our solution has the following preferable properties:
	\textbf{1)} \textbf{Generality}: It can be easily integrated with any 3D segmentation model with minor structural modification;
	\textbf{2)} \textbf{Flexibility}: The fusion module is only used during the training to enhance the 3D network. After training, the enhanced 3D model can be deployed without image inputs.
	\textbf{3)} \textbf{Effectively}: Even with only a small section of overlapped  multi-modality data, our method can significantly boost the performance.
	As a result, we evaluate 2DPASS with a simple yet strong baseline implemented with sparse convolutions~\cite{SparseConv}. The experiments show 2DPASS brings noticeable improvements even over this strong baseline. 
	Equipped with 2DPASS using multi-modal data, our model achieves the \textbf{top-1} results on the single and multiple-scan leaderboards of SemanticKITTI~\cite{behley2019semantickitti}. 
	The state-of-the-art results on the NuScenes~\cite{nuscenes} dataset further confirm the generality of our method.

	In general, the main contributions are summarized as follows.
	\begin{itemize}
		\item We propose {2D Priors Assisted Semantic Segmentation} ({2DPASS}) that assists 3D LiDAR semantic segmentation with 2D priors from cameras. To the best of our knowledge, 2DPASS is the first method that distills multi-modal knowledge to single point cloud modality for semantic segmentation.
		
		\item Equipped with the proposed multi-scale fusion-to-single knowledge distillation (MSFSKS) strategy, 2DPASS achieves the significant performance gains on SemanticKITTI and NuScenes benchmarks, ranking the \textbf{1st} on single and multiple tracks of SemanticKITTI.
	\end{itemize}

	\section{Related Work}
	\subsection{Single-Sensor Methods}
	
	\noindent\textbf{Camera-Based Methods. }
	Camera-based semantic segmentation aims to predict the pixel-wise labels for input 2D images. 
	FCN~\cite{long2015fully} is the pioneer in semantic segmentation, which proposes an end-to-end fully convolutional architecture based on image classification networks. 
	Recent works have achieved significant improvements via exploring multi-scale features learning~\cite{chen2017deeplab,lin2016efficient,zhao2017pyramid}, dilated convolution~\cite{wang2018understanding,chen2017rethinking}, and attention mechanisms\cite{huang2019ccnet,yuan2018ocnet}. 
	However, camera-only methods are ambiguous in depth sensing and not robust in low light conditions.
	
	\noindent\textbf{LiDAR-Based Methods. }
	The LiDAR data is generally represented as point clouds. There are several mainstreams to process point clouds with different representations.
	\textbf{1)} \textbf{Point-based} methods approximate a permutation-invariant set function using a per-point Multi-Layer Perceptron (MLP). PointNet~\cite{qi2017pointnet} is the pioneer in this field. 
	Later on, many studies design point-wise MLP~\cite{qi2017pointnet++,wang2019dynamic}, adaptive weight~\cite{PointConv,liu2019relation} and pseudo grid~\cite{Thomas_2019_ICCV,hua2018pointwise} based methods to extract local features of point clouds or exploit nonlocal operators~\cite{yan2020pointasnl,zhao2021point,engel2021point} to learn long distance dependency.
	However, point-based methods are not efficient in the LiDAR scenario since their sampling and grouping algorithms are generally time-consuming. 
	\textbf{2)} \textbf{Projection-based} methods are very efficient approaches for LiDAR point clouds. They project point clouds onto 2D pixels so that traditional CNN can play a normal role. Previous works project all points scanned by the rotating LiDAR onto 2D images by plane projection~\cite{lawin2017deep, boulch2017unstructured, tatarchenko2018tangent}, spherical projection~\cite{wu2018squeezeseg,wu2019squeezesegv2} or both~\cite{liong2020amvnet}.
	However, the projection inevitably causes information loss. And the projection-based methods currently meet the bottleneck of the segmentation accuracy.
	\textbf{3)} Most recent works adopt \textbf{voxel-based} frameworks since they balance the efficiency and effectiveness, where sparse convolution (SparseConv)~\cite{SparseConv} are most commonly  utilized.
	Compared to traditional voxel-based methods (\ie, 3DCNN) directly transforming all points into the 3D voxel grids, SparseConv only stores non-empty voxels in a Hash table and conducts convolution operations only on these non-empty voxels in a more efficient way. 
	Recently, many studies have used SparseConv to design more powerful network architectures. 
	Cylinder3D~\cite{zhou2020cylinder3d} changes original grid voxels to cylinder ones and designs an asymmetrical network to boost the performance.
	AF$^2$-S3Net~\cite{cheng20212} applies multiple branches with different kernel sizes, aggregating multi-scale features via an attention mechanism.
	\textbf{4)} Very recently, there is a trend of exploiting \textbf{multi-representation fusion} methods. 
	These methods combine multiple representations above (\ie, points, projection images, and voxels) and design feature fusion among different branches.
	Tang~\etal~\cite{tang2020searching} combines point-wise MLPs in each sparse convolution block to learn a point-voxel representation and uses NAS to search for a more efficient architecture.
	RPVNet~\cite{xu2021rpvnet} proposes range-point-voxel fusion network to utilizes information from three representations.
	Nevertheless, these methods only take sparse and textureless LiDAR point clouds as inputs, thus appearance and texture in the camera images have not been fully utilized.
	
	\subsection{Multi-Sensor Methods}
	Multi-sensor methods attempt to fuse information from two complementary sensors and leverage the benefits of both camera and LiDAR~\cite{krispel2020fuseseg,el2019rgb,meyer2019sensor,vora2020pointpainting}.
	RGBAL~\cite{el2019rgb} converts RGB images to a polar-grid mapping representation and designs early and mid-level fusion strategies. 
	PointPainting~\cite{vora2020pointpainting} exploits the segmentation logits of images and projects them to the LiDAR space by bird's-eye projection~\cite{yuan2018ocnet} or spherical projection~\cite{milioto2019rangenet++} for LiDAR network performance improvement. 
	Recently, PMF~\cite{zhuang2021perception} exploits a collaborative fusion of two modalities in camera coordinates.
	However, these methods require multi-sensor inputs in both training and inference phases. 
	Moreover, the paired multi-modality data is usually computation-intensive and unavailable in practical application.
	
	\subsection{Cross-modal Knowledge Transfer}
	
	Knowledge distillation was initially proposed for compressing the large teacher network to a small student one~\cite{hinton2015distilling}.
	Over the past few years, several subsequent studies enhanced knowledge transferring through matching feature representations in different manners~\cite{ba2013deep,chen2017learning,zagoruyko2016paying,srinivas2018knowledge}.
	For instance, aligning attention maps~\cite{zagoruyko2016paying} and Jacobean matrixes~\cite{srinivas2018knowledge} were independently applied.
	With the development of multi-modal computer vision, recent research apply knowledge distillation to transfer priors across different modalities, \eg, exploiting extra 2D images in the training phase and improving the performance in the inference~\cite{gupta2016cross,wang2019efficient,yuan2018rgb,liu20213d,zhao2020knowledge}.
	Specifically, \cite{liu2021learning} introduces the 2D-assisted pre-training, \cite{xu2021image2point} inflates the kernels of 2D convolution to the 3D ones, and \cite{yuan2022x} applies well-designed teacher-student framework.
	Inspired but different from the above, we transfer 2D knowledge through a multi-scale fusion-to-single manner, which additionally takes care of the modal-specific knowledge.

		\begin{figure*}[t]
		\begin{centering}
			\includegraphics[width=0.9\textwidth]{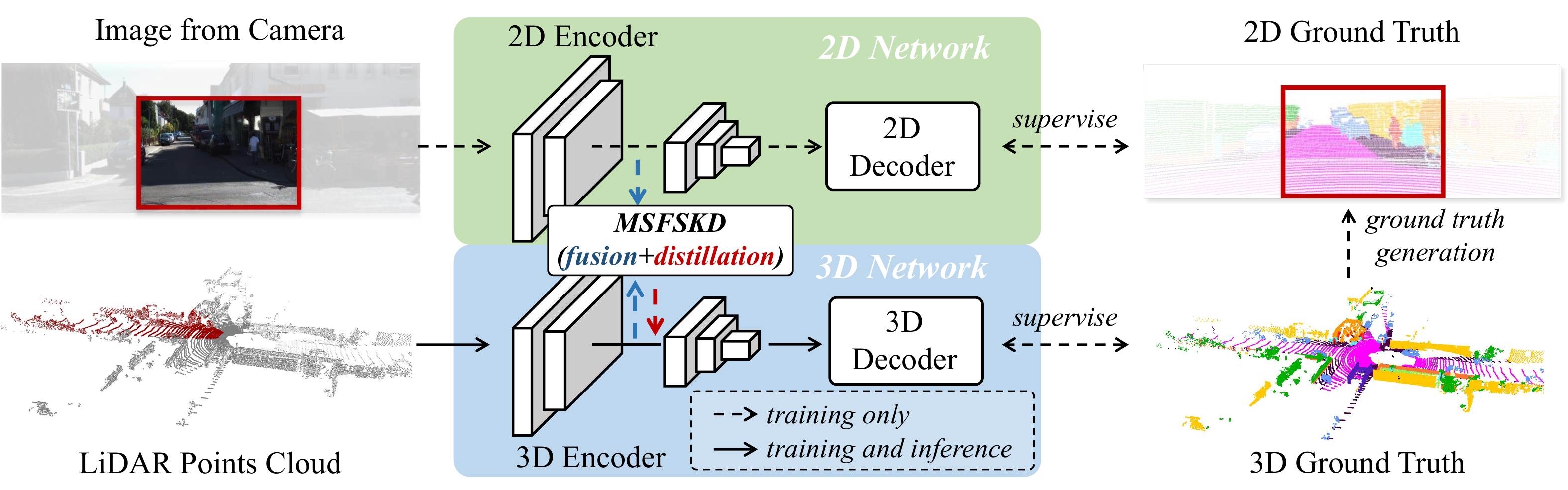}
			\caption{\textbf{{2D Priors Assisted Semantic Segmentation} ({2DPASS}).} It first crops a small patch from the original camera image as the 2D input.
				Then the cropped image patch and LiDAR point cloud independently pass through the 2D and 3D encoders to generate multi-scale features in parallel.
				Afterwards, for each scale, complementary 2D knowledge is effectively transferred to the 3D network via the multi-scale fusion-to-single knowledge distillation (MSFSKD).
				The feature maps (in the form of either pixel grid or point set) are used to generate the final semantic scores using modal-specific decoders, which are supervised by pure 3D labels.
			}
			\label{fig:fig2}
		\end{centering}	
	\end{figure*}

	\section{Method}
	\subsection{Framework Overview}
	This paper focuses on improving the LiDAR point cloud semantic segmentation, which aims to assign the semantic label to each point.
	To handle difficulties in large-scale outdoor LiDAR point clouds, \ie, sparsity, varying density, and lack of texture, we introduce the strong regularization and priors from 2D camera images through a \textbf{fusion-to-single} knowledge transferring.

	The workflow of our {2D Priors Assisted Semantic Segmentation} ({2DPASS}) is shown in Fig.~\ref{fig:fig2}.
	Since the camera images are pretty large (\eg, $1242 \times 512$), sending the original ones to our multi-modal pipeline is intractable.
	Therefore, we randomly sample a small patch ($480 \times 320$) from the original camera image as the 2D input \cite{jaritz2020xmuda}, accelerating the training processing without performance drop.
	Then the cropped image patch and LiDAR point cloud independently pass through independent 2D and 3D encoders, where multi-scale features from the two backbones are extracted in parallel.
	Afterwards, multi-scale fusion-to-single knowledge distillation (MSFSKD) is conducted to enhance the 3D network using multi-modal features, \ie, fully utilizing texture and color-aware 2D priors as well as retaining the original 3D-specific knowledge.
	Finally, all the 2D and 3D features at each scale are used to generate semantic segmentation predictions, which are supervised by pure 3D labels.
	During inference, the 2D-related branch can be discarded, which effectively prevents extra computational burden in real application compared with fusion-based approaches.

	\subsection{Modal-Specific Architectures}
	\label{sec:modal-specific}
	\noindent\textbf{Multi-Scale Feature Encoders.}
	As shown in Fig.~\ref{fig:fig2}, we use two different networks to independently encode multi-scale features from 2D image and 3D point cloud.
	We apply ResNet34~\cite{he2016deep} encoder with 2D convolution as the 2D network.
	For the 3D network, we adopt sparse convolution~\cite{SparseConv} to construct the 3D network.
	One merit of sparse convolution lies in the sparsity, with which the convolution operation only considers the non-empty voxels. 
	Specifically, we design a hierarchical point-voxel encoder as that used in the decoder of \cite{tang2020searching}, and adopt the ResNet bottleneck~\cite{he2016deep} in each scale while replacing the ReLU with Leaky ReLU~\cite{maas2013rectifier}.
	In both network, we extract $L$ feature maps from different scales, obtaining the 2D and 3D features, \ie, $\{F^{2D}_l\}_{l=1}^L$ and $\{F^{3D}_l\}_{l=1}^L$.
	
	\noindent\textbf{Prediction Decoders.}
	After processing the features from images and point clouds at each scale, two modal-specific prediction decoders are independently applied to restore the down-sampled feature maps to their original sizes.

	For the 2D network, we adopt FCN~\cite{long2015fully} decoder to up-sample the features from each encoder layer.
	Specifically, the feature map $D^{2D}_{l}$ from the $l$-th decoder layer can be gained by up-sampling the feature map from the $(L-l+1)$-th encoder layer, where all the up-sampled feature maps will be merged through element-wise addition.
	Finally, the semantic segmentation of the 2D network is obtained by passing the fused feature map through a linear classifier.
	
	For the 3D network, we do not adopt the U-Net decoder used in previous methods~\cite{zhou2020cylinder3d,tang2020searching,cheng20212}.
	In contrast, we up-sample the features from different scales to the original size and concatenate them together before feeding them into the classifier.
	We find out that such a structure can better learn hierarchical information while gaining the prediction in a more efficient way.
	
	\subsection{Point-to-Pixel Correspondence}
	\label{sec:feat}
\begin{figure*}[t]
		\begin{centering}
			\includegraphics[width=\textwidth]{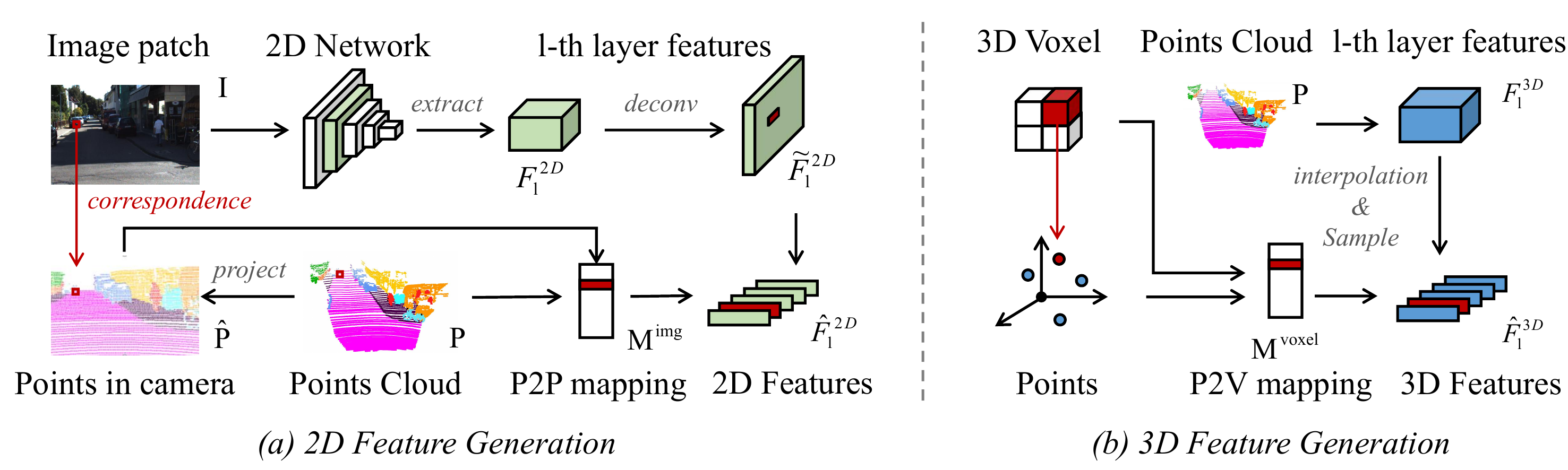}
			\caption{\textbf{2D and 3D feature generation.} 
				Part (a) demonstrates the 2D feature generation, where the point cloud will first be projected onto the image patch and generate the point-to-pixel (P2P) mapping.
				After that, it transfers the 2D feature map to the point-wise 2D features according to P2P mapping.
				Part (b) shows the 3D feature generation.
				The point-to-voxel (P2V) mapping is easy to obtain, and the voxel features will be interpolated onto the point cloud.
			}
			\label{fig:fig3}
		\end{centering}	
	\end{figure*}

	Since the 2D features and 3D features are generally represented as pixels and points, respectively, it is difficult to directly transfer information between two modalities.
	In this section, we aim to generate paired features of two modalities for further knowledge distillation, using the point-to-pixel correspondence.
	The details of paired feature generation in two modalities are demonstrated in Fig.~\ref{fig:fig3}.

	\noindent\textbf{2D Features.}
	The process of 2D feature generation is illustrated in Fig.~\ref{fig:fig3}~(a).
	By cropping a small patch $I \in \mathbb{R}^{H\times W\times 3}$ from the original image and passing it through a 2D network, multi-scale features can be extracted in the hidden layers with different resolution.
	Taking the feature map $F^{2D}_l \in \mathbb{R}^{H_l\times W_l\times D_l} $ from $l$-th layer as an example, we first conduct a decovolution operation to upscale its resolution to the original one $\tilde{F}^{2D}_l$.
	Similar to the recent multi-sensor method~\cite{zhuang2021perception}, we adopt perspective projection and calculate a point-to-pixel mapping between point clouds and images.
	Specifically, given a LiDAR point cloud $P = \{p_i\}_{i=1}^N \in \mathbb{R}^{N\times 3}$, the projection of each 3D point $p_i = (x_i, y_i, z_i) \in \mathbb{R}^{3}$ to a point $\hat{p}_i =(u_i, v_i) \in \mathbb{R}^{2}$ in the image plane is given as:
	\begin{align}
	[u_i, v_i, 1]^T = \frac{1}{z_i} \times K\times T \times [x_i, y_i, z_i, 1]^T,
	\label{project}
	\end{align}
	where $K \in \mathbb{R}^{3\times 4}$ and $T \in \mathbb{R}^{4\times 4}$ are the camera intrinsic and extrinsic matrices respectively.
	$K$ and $T$ are directly provided in KITTI~\cite{geiger2012cvpr}.
	Since the lidar and cameras operate at different frequencies in NuScenes~\cite{nuscenes}, we need to transform the LiDAR frame at timestamp $t_l$ to camera frame at timestamp $t_c$ via the global coordinate system.
	The extrinsic matrix $T$ in NuScenes dataset~\cite{nuscenes} is given as:
	\begin{align}
	T=T_{\text {camera } \leftarrow \mathrm{ego}_{\mathrm{t}_{c}}} \times 
	T_{\mathrm{ego}_{\mathrm{t}_{\mathrm{c}}} \leftarrow \mathrm{global}}\times 
	T_{\mathrm{global} \leftarrow \mathrm{ego}_{\mathrm{t}_{l}}}\times 
	T_{\mathrm{ego}_{\mathrm{t}_{l}} \leftarrow \mathrm{lidar}}
	\label{nuscene_trans}
	\end{align}
	After the projection, the point-to-pixel mapping is represented as 
	\begin{align}
	M^{img} = \{(\lfloor  v_i\rfloor , \lfloor  u_i \rfloor )\}_{i=1}^N  \in \mathbb{R}^{N\times 2},
	\label{map_img}
	\end{align}
	where $\lfloor \cdot \rfloor$ is the floor operation.
	According to the point-to-pixel mapping, we extract a point-wise 2D feature $\hat{F}^{2D} \in \mathbb{R}^{N^{img}\times D_l}$ from the original feature map ${F}^{2D}$ if any pixel on the feature map is included in $M^{img}$.
	Here $N^{img} < N$ represents the number of points that are included in $M^{img}$.
	
	\noindent\textbf{3D Features.}
	The process of 3D features is relatively straightforward (as shown in Fig.~\ref{fig:fig3}~(b)).
	Specifically, for the point cloud $P = \{(x_i, y_i, z_i)\}_{i=1}^N$, we obtain a point-to-voxel mapping in the $l$-th layer through  
	\begin{align}
	M^{voxel}_l = \{(\lfloor  x_i/r_l \rfloor , \lfloor  y_i/r_l \rfloor, \lfloor  z_i/r_l \rfloor )\}_{i=1}^N  \in \mathbb{R}^{N\times 3},
	\label{map_voxel}
	\end{align}
	where $r_l$ is the voxelization resolution in the $l$-th layer.
	After that, given the 3D feature ${F}^{3D}_l \in \mathbb{R}^{N'_l\times D_l}$ from a sparse convolution layer, we gain a point-wise 3D feature $\tilde{F}^{3D}_l \in \mathbb{R}^{N\times D_l}$ through nearest interpolation on the original feature map ${F}^{3D}_l$ according to $M^{voxel}_l$.
	Finally, we filter the points by discarding points outside the image FOV:
	\begin{align}
	\hat{F}^{3D}_l = \{f_i | f_i \in \tilde{F}^{3D}_l, M_{i,1}^{img} \leq H, M_{i,2}^{img} \leq W\}_{i=1}^N  \in \mathbb{R}^{N^{img}\times D_l},
	\label{keep_img}
	\end{align}

	\noindent\textbf{2D Ground Truths.} Considering only 2D images is provided, the 2D ground-truths are obtained by projecting the 3D point labels to the corresponding image plane using above point-to-pixel mapping.
	Afterwards, the projected 2D ground truths can work as the supervision for the 2D branch.

	\noindent\textbf{Features Correspondence.}
	Since both 2D and 3D feature use the same point-to-pixel mapping, 2D features $\hat{F}^{2D}_l$ and 3D features $\hat{F}^{3D}_l$ in arbitrary $l$-th layer have the same number of point $N^{img}$ and point-to-pixel correspondence.

	\begin{figure*}[t]
		\begin{centering}
			\includegraphics[width=0.9\textwidth]{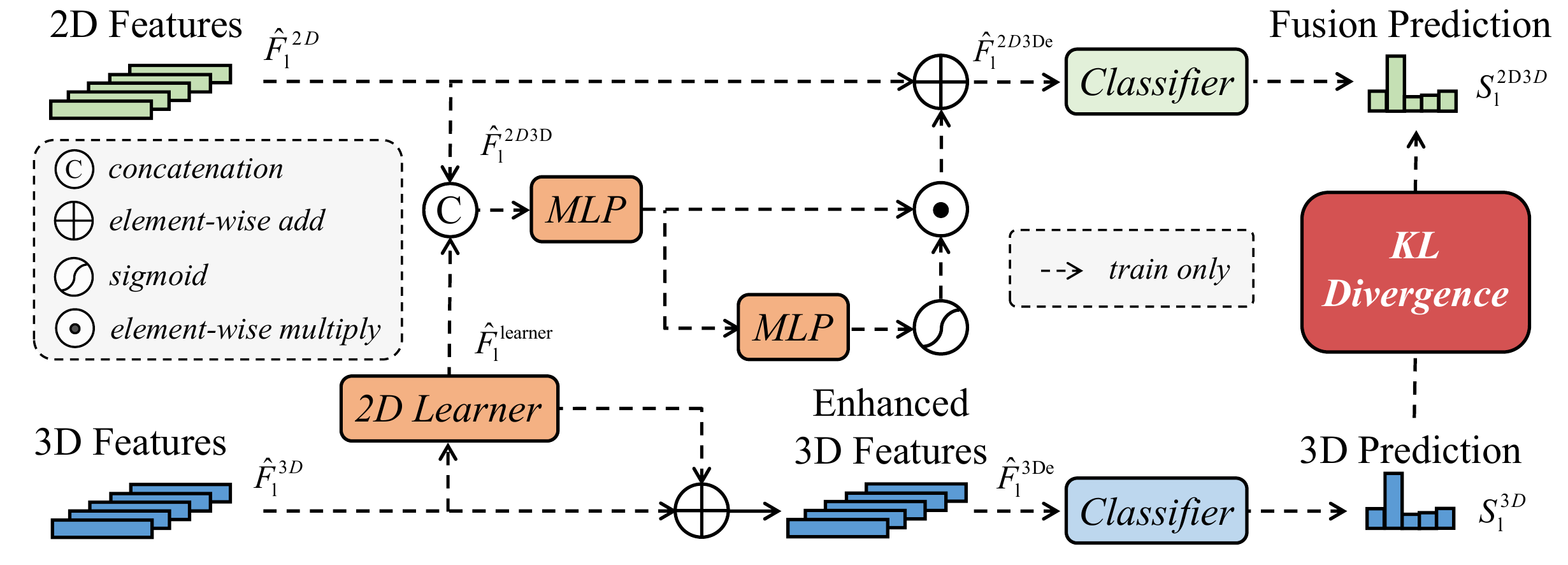}
			\caption{Internal structure of \textbf{Multi-Scale Fusion-to-Single Knowledge Distillation (MSFSKD)}, which consists of the modality fusion and Modality-Preserving KD. For each scale, modality fusion is first ultilized to achieve an enhanced multi-modality feature $\hat{F}^{2D3D_e}_l$. Afterwards, the enhanced feature $\hat{F}^{2D3D_e}_l$ promotes the 3D representation $\hat{F}^{3D_{e}}_l$ through the uni-directional Modality-Preserving KD.
			}
			\label{fig:fig4}
		\end{centering}	
	\end{figure*}

	\subsection{Multi-Scale Fusion-to-Single Knowledge Distillation (MSFSKD)}
	\label{sec:ms}
	As the key of 2DPASS, MSFSKD aims at improving the 3D representation in each scale using auxiliary 2D priors through a fusion-then-distillation manner.
	The knowledge distillation (KD) design of MSFSKD is partially inspired by \cite{jaritz2020xmuda}.
	However, \cite{jaritz2020xmuda} conducts KD in a naive cross-modal manner, \ie, simply aligning the outputs from two sets of single modal features (\ie~ either 2D or 3D), which inevitably pushes the features from two modals to their overlapped space.
	Therefore, such a manner actually discards the modal-specific information, which is crucial in multi-sensor segmentation.
	Although this issue can be relieved by introducing extra segmentation heads~\cite{jaritz2020xmuda}, it is inherent for the cross-modal distillation, resulting in biased predictions.
	To this end, we propose multi-scale fusion-to-single knowledge distillation (MSFSKD) module as shown in Fig.~\ref{fig:fig4}, which first fuses features of both images and point clouds and then conducts unidirectional alignment between the fused and the point cloud features.
	In our fusion-then-distillation manner, the fusion well retains the complete information from multi-modal data.
	Besides, the unidirectional alignment ensures boosted point cloud features from fusion without losing modal-specific information.

    \noindent\textbf{Modality Fusion.}
	For each scale, considering the 2D and 3D feature gaps owing to different backbones, it is ineffective to directly fuse the raw 3D features $\hat{F}^{3D}_l$ into their 2D counterparts $\hat{F}^{2D}_l$.
	Thus, we firstly transform $\hat{F}^{3D}_l$ to $\hat{F}^{{{\text{learner}}}}_l$ through a ``2D learner" MLP, which struggles to narrow the feature gap.
	Afterwards, the $\hat{F}^{{{\text{learner}}}}_l$ not only flows into the subsequent concatenation with 2D features $\hat{F}^{2D}_l$ to gain the fused features $\hat{F}^{2D3D}_l$ through another MLP, but also goes back into the original 3D features via a skip connection to yield enhanced 3D features $\hat{F}^{3D_{e}}_l$.
	Besides, similar to attention mechanism, the final enhanced fused features $\hat{F}^{2D3D_e}_l$ is obtained by:
	\begin{align}
	\hat{F}^{2D3D_e}_l = \hat{F}^{2D}_l +& \sigma(\texttt{MLP}(\hat{F}^{2D3D}_l)) \odot  \hat{F}^{2D3D}_l,
	\label{fuse}
	\end{align}
	where $\sigma$ denotes Sigmoid activation function.
	
	\noindent\textbf{Modality-Preserving KD.} Although the $\hat{F}^{{{\text{learner}}}}_l$ is generated from pure 3D features, it is influenced by the segmentation loss of the 2D decoder as well, which takes enhanced fused feature $\hat{F}^{2D3D_e}_l$ as inputs.
	Acting like a residual between fused and point features, the 2D learner feature $\hat{F}^{{{\text{learner}}}}_l$ well prevents the distillation from contaminating the modal-specific information in $\hat{F}^{3D}_l$, achieving a Modality-Preserving KD.
	%
	Finally, two independent classifiers (fully-connected layers) are respectively applied on top of $\hat{F}^{2D3D_e}_l$ and $\hat{F}^{3D_{e}}_l$ to obtain the semantic scores $S^{2D3D}_l$ and $S^{3D}_l$.
	We choose KL divergence as the distillation loss $L_{xM}$ as follows:
	\begin{align}
	L_{xM} &= D_{KL}(S^{2D3D}_l||S^{3D}_l).
	\label{kl}
	\end{align}
	Through such an implementation, it enforces the uni-directional distillation by pushing $S^{3D}_{l}$ closer to $S^{2D3D}_l$.

	By taking such a knowledge distillation scheme, there are several advantages in our framework:
	\textbf{1)} The 2D learner and the fusion-to-single distillation provides rich texture information and structural regularization to enhance the 3D feature learning without losing any modal-specific information in 3D.
	\textbf{2)} The fusion branch is only adopted in the training phase. Therefore, the enhanced model can almost run without extra computational cost during the inference.

	\begin{table*}[t]
		\small
		\renewcommand\tabcolsep{1.5pt} 
		\caption{Semantic segmentation results on the \textit{SemanticKITTI} test benchmark. Only approaches published before 03/08/2022 are compared.}
		
		\begin{center}
			{\fontfamily{cmr}
				\resizebox{\textwidth}{!}{
					\begin{tabular}{lc|ccccccccccccccccccc|c}
						\hline
						
						Method& 
						\rotatebox{90}{mIoU}&
						\rotatebox{90}{road}&
						\rotatebox{90}{sidewalk}&
						\rotatebox{90}{parking}&
						\rotatebox{90}{other-ground~}&
						\rotatebox{90}{building}&
						\rotatebox{90}{car}&
						\rotatebox{90}{truck}&
						\rotatebox{90}{bicycle}&
						\rotatebox{90}{motorcycle}&
						\rotatebox{90}{other-vehicle}&
						\rotatebox{90}{vegetation}&
						\rotatebox{90}{trunk}&
						\rotatebox{90}{terrain}&
						\rotatebox{90}{person}&
						\rotatebox{90}{bicyclist}&
						\rotatebox{90}{motorcyclist}&
						\rotatebox{90}{fence}&
						\rotatebox{90}{pole}&
						\rotatebox{90}{traffic sign} &
						\rotatebox{90}{speed (ms)} \\
						\hline
						\hline

						SqueezeSegV2~\cite{wu2019squeezesegv2}  &39.7&88.6& 67.6& 45.8& 17.7& 73.7& 81.8& 13.4& 18.5& 17.9& 14.0& 71.8& 35.8 &60.2& 20.1& 25.1& 3.9& 41.1& 20.2& 26.3 & -\\
						
						DarkNet53Seg~\cite{behley2019semantickitti} &49.9 &91.8 &74.6 &64.8 &27.9 &84.1 &86.4 &25.5 &24.5 &32.7 &22.6 &78.3 &50.1 &64.0 &36.2 &33.6 &4.7 &55.0 &38.9 &52.2 & - \\
						
						RangeNet53++~\cite{milioto2019rangenet++} &52.2 &91.8 &75.2 &65.0 &27.8 &87.4 &91.4 &25.7 &25.7 &34.4 &23.0 &80.5 &55.1 &64.6 &38.3 &38.8 &4.8 &58.6 &47.9 &55.9 & 83.3\\
						
						3D-MiniNet~\cite{alonso20203d} &55.8 &91.6 &74.5 &64.2 &25.4 &89.4 &90.5 &28.5 &42.3 &42.1 &29.4 &82.8 &60.8 &66.7 &47.8 &44.1 &14.5 &60.8 &48.0 &56.6 & -\\
						SqueezeSegV3~\cite{xu2020squeezesegv3} &55.9 &91.7 &74.8 &63.4 &26.4 &89.0 &92.5 &29.6 &38.7 &36.5 &33.0 &82.0 &58.7 &65.4 &45.6 &46.2 &20.1 &59.4 &49.6 &58.9  & 238\\

						PointNet++~\cite{qi2017pointnet++}&20.1& 72.0 &41.8 &18.7& 5.6 &62.3 &53.7 &0.9 &1.9 &0.2& 0.2& 46.5 &13.8 &30.0 &0.9 &1.0 &0.0 &16.9 &6.0 &8.9 &5900\\
						TangentConv~\cite{tatarchenko2018tangent} &40.9 &83.9 &63.9 &{33.4} &{15.4} &{83.4} &{90.8} &15.2&{2.7}& 16.5 &12.1 &79.5 &49.3 &58.1 &23.0 &28.4 &{8.1} &{49.0} &35.8 &28.5 &3000\\
						PointASNL~\cite{yan2020pointasnl}  & 46.8&87.4 &74.3&24.3&1.8&83.1&87.9&39.0&0.0&25.1&29.2&84.1&52.2&70.6&34.2& 57.6&0.0&43.9&57.8&36.9 & -\\
						RandLA-Net~\cite{hu2019randla} &55.9 &90.5 &74.0 &61.8 &24.5 &89.7 &94.2 &43.9 &29.8 &32.2 &{39.1} &83.8 &63.6 &68.6 &48.4 &47.4 &9.4 &60.4 &51.0 &50.7 & 880\\
						KPConv~\cite{Thomas_2019_ICCV} &58.8 &{90.3} &72.7 &{61.3} &31.5 &90.5 &95.0 &33.4 &30.2 &42.5 &44.3 &84.8 &69.2 &69.1 &61.5 &61.6 &11.8 &64.2 &56.4 &47.4 &-\\

						PolarNet~\cite{zhang2020polarnet}&54.3 &90.8 &74.4 &61.7 &21.7 &90.0 &93.8 &22.9 &40.3 &30.1 &28.5 &84.0 &65.5 &67.8 &43.2 &40.2 &5.6 &61.3 &51.8 &57.5 & \bf{62}\\
						
						JS3C-Net~\cite{yan2021sparse}   & 66.0 & 88.9& 72.1& 61.9& 31.9& 92.5& 95.8& 54.3 & 59.3& 52.9& 46.0& 84.5 & 69.8 & 67.9& 69.5 & 65.4 & 39.9 & 70.8 & 60.7 & 68.7 & 471\\
						SPVNAS~\cite{tang2020searching}  & 67.0  & 90.2  & 75.4  & 67.6  & 21.8  & 91.6  & 97.2  & 56.6  & 50.6  & 50.4  & 58.0  & 86.1  & 73.4  & 71.0  & 67.4  & 67.1  & 50.3  & 66.9  & 64.3  & 67.3 &259 \\
						Cylinder3D~\cite{zhou2020cylinder3d}  & 68.9  & 92.2  & 77.0  & 65.0  & 32.3  & 90.7  & 97.1  & 50.8  & 67.6  & 63.8  & 58.5  & 85.6  & 72.5  & 69.8  & 73.7  & 69.2  & 48.0  & 66.5  & 62.4  & 66.2 & 131 \\
						RPVNet~\cite{xu2021rpvnet}  & 70.3  & \bf{93.4}  & \bf{80.7}  & \bf{70.3}  & 33.3  & \bf{93.5}  & \bf{97.6}  & 44.2  & \bf{68.4}  & 68.7  & 61.1  & \bf{86.5}  & \bf{75.1}  & \bf{71.7}  & 75.9  & 74.4  & 43.4  & 72.1  & 64.8  & 61.4  & 168 \\
						(AF)$^2$-S3Net~\cite{cheng20212}  & 70.8  & 92.0  & 76.2  & 66.8  & \textbf{45.8}  & 92.5  & 94.3  & 40.2  & 63.0  & \bf{81.4}  & 40.0  & 78.6  & 68.0  & 63.1  & 76.4  & \bf{81.7}  & \bf{77.7}  & 69.6  & 64.0  & \bf{73.3} &- \\
						\hline
						Baseline & 67.4  & 89.8  & 73.8  & 62.1  & 33.5  & 91.9  & 96.3  & 54.9  & 51.1  & 55.8  & 51.6  & \bf{86.5}  & 72.3  & 71.3  & 76.8  & 79.8  & 30.3  & 68.7  & 63.7  & 70.2 & \bf{62}\\
						\textbf{2DPASS(Ours)}  & \textbf{72.9}  & 89.7  & 74.7  & 67.4  & 40.0  & \textbf{93.5}  & 97.0  & \textbf{61.1}  & 63.6  & 63.4  & \bf{61.5}  & 86.2  & 73.9  & 71.0  & \bf{77.9}  & 81.3  & 74.1  & \bf{72.9}  & \bf{65.0} &70.4 & \bf{62} \\\hline
					\end{tabular} 
			}}
		\end{center}
		\label{tab:kitti_seg}
	\end{table*}

	\section{Experiments}
	\subsection{Experiment Setups}
	\noindent\textbf{Datasets. }
	We extensively evaluate 2DPASS on two large-scale outdoor benchmarks: SemanticKITTI~\cite{behley2019semantickitti} and Nuscenes~\cite{nuscenes}. \textbf{SemanticKITTI} provides dense semantic annotations for each individual scan of sequences 00-10 in KITTI dataset~\cite{geiger2012cvpr}. 
	According to the official setting, sequence 08 is the validation split, while the remaining are the train split.
	SemanticKITTI uses sequences 11-21 in KITTI as the test set, whose labels are held on for blind online testing\footnote{\url{https://competitions.codalab.org/competitions/20331}}.
	\textbf{NuScenes} contains 1000 scenes which show a great diversity in inner cities traffic and weather conditions. It officially divides the data into 700/150/150 scenes for train/val/test. 
	Similar to SemanticKITTI, the test set of NuScenes is used for online benchmarking\footnote{\url{https://eval.ai/web/challenges/challenge-page/720/leaderboard/1967}}.
	For 2D sensors, KITTI has only two front-view cameras, while NuScenes has six cameras covering the full 360\degree~fields of view.
	
	\noindent\textbf{Evaluation Metrics.} 
	We evaluate methods mainly using mean intersection over union (mIoU), which is defined as the average IoU over all classes. 
	Additionally, we report the overall accuracy (Acc)/ frequency-weighted IOU (FwIOU) provided by the online leaderboard of two benchmarks.
	FwIoU is similar to mIoU except that each IoU is weighted by the point-level frequency of its class.
	
	\noindent\textbf{Network Setup.} 
	We apply ResNet34~\cite{he2016deep} encoder with 2D convolution as the 2D network, where features after each down-sampling layers are extracted to generate 2D features.
	The 3D encoder is a modified SPVCNN~\cite{tang2020searching} (voxel size 0.1) with fewer parameters, whose hidden dimensions are 64 for SemanticKITTI and 128 for NuScenes to speed up the network.
	The number of layers $L$ for MSFSKD is set to 4 and 6 for SemanticKITTI and NuScenes, respectively.
	In each scale of knowledge distillation, 2D and 3D features are reduced to 64 dimensions through deconvolution or MLPs.
	Similarly, the hidden size of MLPs and 2D learner in MSFSKD are identically 64.
	
	\noindent\textbf{Training and Inference Details.} 
	We employ the cross-entropy and Lovasz losses as \cite{zhou2020cylinder3d} for semantic segmentation. 
	For the knowledge distillation, we set the proportion of segmentation loss and KL divergence as $1:0.05$.
	Test-time augmentation~\cite{zhou2020cylinder3d} is applied during the inference.
	Training details will be introduced in supplementary material.

	\begin{table*}[t]		
		\small
		\renewcommand\tabcolsep{1.5pt} 
		\caption{Comparison to the state-of-the-art methods on the test set of SemanticKITTI multiple scans challenge. \textit{-s} indicates static and \textit{-m} stands for moving.}
		\begin{center}
			{\fontfamily{cmr}
				\resizebox{0.85\textwidth}{!}{%
					\begin{tabular}{l|cc|cccccccccccc}
						\hline 
						Method & 
						\rotatebox{90}{\bf{mIoU}}&
						\rotatebox{90}{\bf{Acc}}&
						
						\rotatebox{90}{car-s}&
						\rotatebox{90}{car-m}&
						\rotatebox{90}{truck-s}&
						\rotatebox{90}{truck-m}&
						\rotatebox{90}{other-s}&
						\rotatebox{90}{other-m~}&
						\rotatebox{90}{person-s}&
						\rotatebox{90}{person-m}&
						\rotatebox{90}{bicyclist-s}&
						\rotatebox{90}{bicyclist-m}&
						\rotatebox{90}{motorcyclist-s}&
						\rotatebox{90}{motorcyclist-m~} \\
						\hline 
						\hline
						
						LatticeNet~\cite{rosu2019latticenet} & 45.2  & 89.3  & 91.1  & 54.8  & 29.7  & 3.5   & 23.1  & 0.6   & 6.8   & 49.9  & 0.0   & 44.6  & 0.0   & 64.3  \\
						TemporalLidarSeg~\cite{duerr2020lidar} & 47.0  & 89.6  & 92.1  & 68.2  & 39.2  & 2.1   & 35.0  & \bf{12.4}  & 14.4  & 40.4  & 0.0   & 42.8  & 0.0   & 12.9  \\
						KPConv~\cite{Thomas_2019_ICCV} & 51.2  & 89.3  & 93.7  & 69.4  & 42.5  & 5.8   & 38.6  & 4.7   & 21.6  & 67.5  & 0.0   & 67.4  & 0.0   & 47.2  \\
						Cylinder3D~\cite{zhou2020cylinder3d} & 52.5  & 91.0  & 94.6  & 74.9  & 41.3  & 0.0   & 38.8  & 0.1   & 12.5  & 65.7  & 1.7   & 68.3  & 0.2   & 11.9  \\
						(AF)$^2$-S3Net~\cite{cheng20212} & 56.9  & 88.1  & 91.8  & 65.3  & 15.7  & 5.6   & 27.5  & 3.9   & 16.4  & 67.6  & \bf{15.1}  & 66.4  & \bf{67.1}  & 59.6  \\
						\hline 
						\textbf{2DPASS(Ours)} &  \bf{62.4}  &  \bf{91.4}  & \bf{96.2}  &\bf{82.1}  & \bf{48.2}  & \bf{16.1}  & \bf{52.7}  & 3.8   & \bf{35.4}  & \bf{80.3}  & 7.9   & \bf{71.2}  & 62.0  & \bf{73.1}  \\
						\hline
			\end{tabular}}}
		\end{center}
		\label{tab:kitti-multi}
	\end{table*}
	
	\begin{figure*}[t]
		\begin{centering}
			\includegraphics[width=\textwidth]{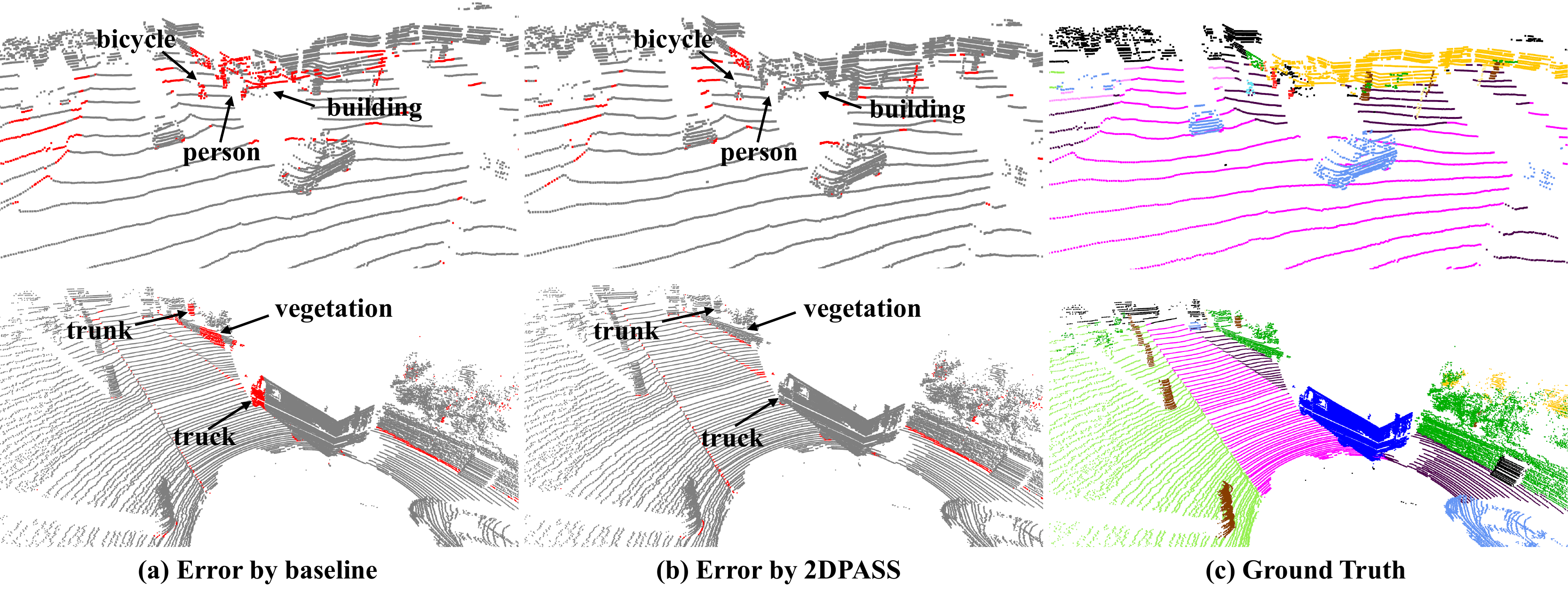}
			\caption{{Qualitative results of 2DPASS on the validation set of SemanticKITTI. Our baseline has a higher error recognizing small objects and region boundaries, while 2DPASS recognizes small objects better thanks to the prior of 2D modality.} 
			}
			\label{fig:fig_vis_kitti}
		\end{centering}	
	\end{figure*}

	\subsection{Benchmark Results}
	\noindent\textbf{SemanticKITTI.}
	SemanticKITTI evaluates segmentation performance using two settings: single scan and multiple scans.
	For methods using a single scan as input, moving and non-moving are mapped to a single class. While methods using multiple scans as inputs should distinguish between moving and non-moving objects, which is more challenging.
	All the reported results are from the official blind test competition website of SemanticKITTI.
	
	Tab.~\ref{tab:kitti_seg} shows our performance under the single scan setting.
	Our baseline without 2DPASS already performs on par with a strong model Cylinder3D~\cite{zhou2020cylinder3d} while runs at a faster speed.
	Even so, the application of 2DPASS still brings a significant improvement over the baseline.
	Thanks to the auxiliary knowledge distillation,  2DPASS does not put any extra burden on the original model and thus does not sacrifice the running speed of the baseline.
	Overall, 2DPASS achieves the best result in terms of mIoU and running speed, outperforming the state-of-the-art (\ie,  (AF)$^2$-S3Net~\cite{cheng20212}) by {\textbf{2.1\%}}.
	The visualization results on SemanticKITTI single scan are shown in Fig.~\ref{fig:fig_vis_kitti}.
	
	\begin{table*}[t]
		\small
		\renewcommand\tabcolsep{1.5pt} 
		\caption{Semantic segmentation results on the \textit{Nuscenes} test benchmark. Only approaches published before 03/08/2022 are compared. \textit{L} and \textit{C} stand for LiDAR and camera, respectively. (*) The speed reported in PMF~\cite{zhuang2021perception} is accelerated by TensorRT, and we test their model without such technique in the same environment. 
}
		
		\begin{center}
			{\fontfamily{cmr}
				\resizebox{\textwidth}{!}{%
					\begin{tabular}{l|c|cc|cccccccccccccccc|c}
						\hline
						
						Method& Input&
						\rotatebox{90}{\bf{mIoU}}&
						\rotatebox{90}{\bf{FW mIoU}}&
						\rotatebox{90}{barrier}&
						\rotatebox{90}{bicycle}&
						\rotatebox{90}{bus}&
						\rotatebox{90}{car}&
						\rotatebox{90}{construction}&
						\rotatebox{90}{motorcycle}&
						\rotatebox{90}{pedestrian}&
						\rotatebox{90}{traffic cone}&
						\rotatebox{90}{trailer}&
						\rotatebox{90}{truck}&
						\rotatebox{90}{driveable}&
						\rotatebox{90}{other flat}&
						\rotatebox{90}{sidewalk}&
						\rotatebox{90}{terrain}&
						\rotatebox{90}{manmade}&
						\rotatebox{90}{vegetation}&
						\rotatebox{90}{speed (ms)} \\
						\hline
						\hline
						
						\text{PolarNet}~\cite{zhang2020polarnet} & L & 69.4 & 87.4 & 72.2 & 16.8 & 77.0 & 86.5 & 51.1 & 69.7 & 64.8 & 54.1 & 69.7 & 63.5 & 96.6 & 67.1 & 77.7 & 72.1 & 87.1 & 84.5 & - \\
						\text{JS3C-Net}~\cite{yan2021sparse} & L & 73.6 & 88.1 & 80.1 & 26.2 & 87.8 & 84.5 & 55.2 & 72.6 & 71.3 & 66.3 & 76.8 & 71.2 & 96.8 & 64.5 & 76.9 & 74.1 & 87.5 & 86.1 & - \\
						\text{Cylinder3D}~\cite{zhou2020cylinder3d} & L & 77.2 & 89.9 & 82.8 & 29.8 & 84.3 & 89.4 & 63.0 & 79.3 & 77.2 & 73.4 & 84.6 & 69.1 & \bf{97.7} & \bf{70.2} & \bf{80.3} & 75.5 & 90.4 & 87.6  & 63\\
						\text{AMVNet}~\cite{liong2020amvnet} & L & 77.3 & 90.1 & 80.6 & 32.0 & 81.7 & 88.9 & 67.1 & 84.3 & 76.1 & 73.5 & \bf{84.9} & 67.3 & 97.5 & 67.4 & 79.4 & 75.5 & 91.5 & 88.7  &85\\
						\text{SPVCNN}~\cite{tang2020searching} & L &77.4 & 89.7 & 80.0 & 30.0 & 91.9 & 90.8 & 64.7 & 79.0 & 75.6 & 70.9 & 81.0 & 74.6 & 97.4 & 69.2 & 80.0 & 76.1 & 89.3 & 87.1 & 63\\
						\text{(AF)$^2$-S3Net}~\cite{cheng20212} & L & 78.3 & 88.5 & 78.9 & 52.2 & 89.9 & 84.2 & \bf{77.4} & 74.3 & 77.3 & 72.0 & 83.9 & 73.8 & 97.1 & 66.5 & 77.5 & 74.0 & 87.7 & 86.8 &270 \\
						\text {PMF}~\cite{zhuang2021perception}  & L+C &77.0 & 89.0 & 82.0 & 40.0 & 81.0 & 88.0 & 64.0 & 79.0 & 80.0 & \bf{76.0} & 81.0 & 67.0 & 97.0 & 68.0 & 78.0 & 74.0 & 90.0 & 88.0 & 125*\\
						\text {2D3DNet}~\cite{genova2021learning} & L+C &80.0 & \bf{90.1} & \bf{83.0} & \bf{59.4} & 88.0 & 85.1 & 63.7 & 84.4 & \bf{82.0} & \bf{76.0} & 84.8 & 71.9 & 96.9 & 67.4 & 79.8 & \bf{76.0} & \bf{92.1} & \bf{89.2} & - \\
						\hline
						\text{Baseline} & L & 77.6  & 88.5  & 80.8  & 37.9  & 92.7  & 90.5  & 65.4  & 77.6  & 71.5  & 70.9  & 83.1  & 75.3  & 97.0    & 69.3  & 78.1  & 75.6  & 89.1  & 86.8 & \bf{44} \\
						\text {\bf{2DPASS(Ours)}} & L &\bf{80.8} & \bf{90.1} & 81.7 & 55.3 & \bf{92.0} & \bf{91.8} & 73.3 & \bf{86.5} & 78.5 & 72.5 & 84.7 & \bf{75.5} & 97.6 & 69.1 & 79.9 & 75.5 & 90.2 & 88.0 &\bf{44} \\\hline
			\end{tabular} }}
		\end{center}
		\label{tab:nus_seg}
	\end{table*}

	Tab.~\ref{tab:kitti-multi} reports the results under the multiple scans setting.
	The mIoU and overall accuracy are calculated over all 25 classes.
	Due to the limited space, we only report the per-class IOUs for dynamic objects with non-moving/moving properties.
	Under this challenge setting, 2DPASS surprisingly surpasses previous approaches with even larger margins, \ie, achieving better mIoU ({{5.5\% improvement}} over (AF)$^2$-S3Net~\cite{cheng20212}) and overall accuracy.

	\noindent\textbf{NuScenes.}
	The results on NuScenes are reported in Tab.~\ref{tab:nus_seg}, where 2DPASS achieves the 1st place as well.
	Note that we only include published works in Tab.~\ref{tab:nus_seg} and the results are directly taken from the official leaderboard of NuScenes, where our model also ranks the 3rd place with slight disadvantage when considering unpublished works.
	Besides surpassing all single-modal methods, 2DPASS surprisingly outperforms those fusion-based approaches (the last two rows in Tab.~\ref{tab:nus_seg}). 
	Note that NuScenes provides images covering the whole FOV of the LiDAR, and fusion-based approaches achieve such results by using both point clouds and image features during the inference.
	In contrast, our method only takes point clouds as input.

	\begin{table*}[t]

	\begin{floatrow}
		\small
		\capbtabbox{
			\resizebox{0.35\textwidth}{!}{
				{\fontfamily{cmr}	
								\begin{tabular}{l|c}
						\hline
						Method &SemanticKITTI\\
						\hline
						Hinton \etal~\cite{hinton2015distilling} & 66.34  \\
						Huang \etal~\cite{huang2021revisiting} & 66.46 \\
						Yang \etal~\cite{yang2021knowledge} & 66.75  \\
						xMUDA~\cite{jaritz2020xmuda} & 67.88  \\\hline
						2DPASS  & \textbf{69.32}  \\
						\hline
					\end{tabular}%
				}
			}
		}{	
			\caption{Comparison with different knowledge distillation.\vspace{-.08cm}}
			\label{tab:ablation1}
		}
		\small
	
		\capbtabbox{
			\resizebox{0.55\textwidth }{1.05cm}{
				{\fontfamily{cmr}
						\begin{tabular}{c|ccc|c}
				\hline
				\multicolumn{1}{r|}{\multirow{1}[4]{*}{baseline}} & \multicolumn{3}{c|}{MSFSKD} & \multicolumn{1}{r}{\multirow{1}[4]{*}{SemanticKITTI}} \\
				
				& KL Div & Modality Fusion & 2D Learner &  \\
				\hline
				\cmark &             &       &       & 65.58 \\

				\cmark &  \cmark &       &       & 66.34 \\
				
				\cmark &  \cmark & \cmark &       & 69.13 \\
				
				\cmark &  \cmark & \cmark & \cmark & \textbf{69.32} \\
				\hline
			\end{tabular}%
				}
				
			}
		}{
			\caption{Ablation study on the SemanticKITTI validation set.}
			\label{tab:ablation}
		}
	\end{floatrow}
	
\end{table*}

	\subsection{Comprehensive Analysis}

	\noindent\textbf{Comparing with Other Knowledge Distillation.}
	To further verify the effectiveness of our fusion-to-single knowledge distillation paradigm upon common teach-student architecture and other cross-modal manners, we compare 2DPASS with typical approaches of knowledge transfer in Tab.~\ref{tab:ablation1}, where we utilize these methods in each scale for fair comparison.
	Among all the methods, Hinton \etal~\cite{hinton2015distilling}, Huang \etal~\cite{huang2021revisiting} and Yang \etal~\cite{yang2021knowledge} are pure knowledge distillation designs, where the former is the pioneer for the research field and the latter is newly proposed.
	As shown in the Tab.~\ref{tab:ablation1}, pure knowledge distillation manners cannot be directly adopted on the LiDAR semantic segmentation, and their improvement upon the baseline model is limited.
	Recently, \cite{jaritz2020xmuda} adopts cross-modal feature alignment technique in the task of domain adaptation on semantic segmentation.
	However, their improvement is still marginal.
	To the end, in the Tab.~\ref{tab:ablation1}, 2DPASS significantly performs better, which illustrates the effectiveness of our multi-scale fusion-to-single knowledge distillation (MSFSKD).
	
	\noindent\textbf{Design Analysis of MSFSKD.}
	Tab.~\ref{tab:ablation} demonstrates the ablation study on SemanticKITTI validation set.
	As shown in the table, our baseline only achieves a lower result of 65.58 mIoU.
	Note that simply using feature alignment between two modalities cannot effectively improve the result, where the metric of mIoU will be only increased to 66.34.
	After using 2D-3D fusion in each knowledge distillation scale, there is a significant improvement to 69.13.
	This improvement mainly comes from the knowledge provided by the stronger fusion prediction.
	Finally, we find out that 2D learner design can slightly improve the performance by about 0.2\%.
	Note that the results on SemanticKITTI validation set is lower than that on benchmark since small object category (\ie, motocyclist) only occupies a small proportion.

\begin{figure*}[t]
		\begin{centering}
			
			\includegraphics[width=\textwidth]{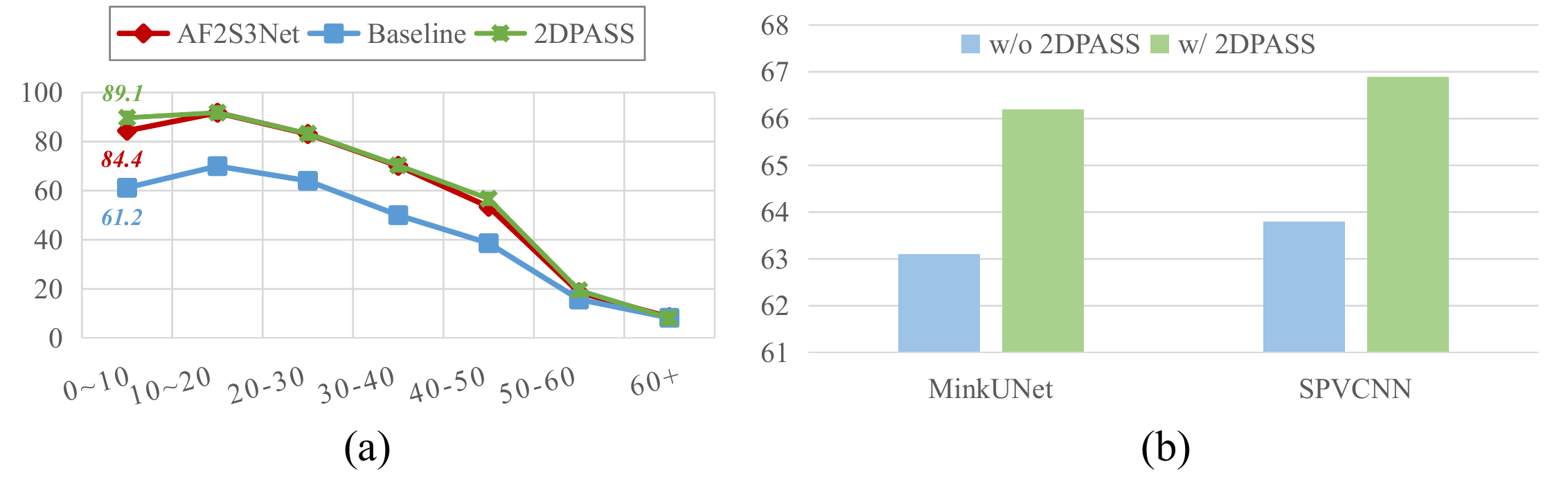}
			\caption{\textbf{Extensive experiment results.} The part (a) shows the results on SemanticKITTI validation set with different distance-range. Part (b) demonstrates the results before and after exploiting 2DPASS on  MinkowskiNet~\cite{tang2020searching} and SPVCNN~\cite{tang2020searching}.
			}
			\label{fig:fig7}
		\end{centering}	
	\end{figure*}

	\noindent\textbf{Distance-based Evaluation.}
	We investigate how segmentation is affected by distance of the points to the ego-vehicle, and compare 2DPASS, current state-of-the-art and the baseline on the SemanticKITTI validation set.
	Fig.~\ref{fig:fig7}~(a) illustrates the mIoU of 2DPASS as opposed to the baseline and (AF)$^2$-S3Net.
	The results of all the methods get worse by increasing the distance since points are relatively sparse in the long distance. 
	2DPASS improves the performance greatly within 10$m$, \ie, from 61.2 to 89.1, which is the best distance for the camera to capture objects' color and texture.
	There is also a significant improvement upon (AF)$^2$-S3Net within this distance, \ie, 84.4 v.s. 89.1.
	
	\noindent\textbf{Generality.}
	We show our 2DPASS can be a ``model-independent'' training scheme that boosts the performance of other networks.
	We additionally trained two open-sourced baselines, \ie, MinkowskiNet and SPVCNN implemented in \cite{tang2020searching} with 2DPASS. During the experiment, we keep all the setups the same except for the 2D-related components.
	As shown in Fig.~\ref{fig:fig7}~(b), 2DPASS improves the former one from 63.1 to 66.2 and the latter from 63.8 to 66.9.
	These results sufficiently demonstrate the effectiveness and generality of 2DPASS.

	\section{Conclusion}
	
	This work proposes the {2D Priors Assisted Semantic Segmentation} ({\textbf{2DPASS}}), a general training scheme, to boost the performance of LiDAR point cloud semantic segmentation via 2D prior-related knowledge distillation.
	By leveraging an auxiliary modal fusion and knowledge distillation in a multi-scale manner, 2DPASS acquires richer semantic and structural information from the multi-modal data, effectively enhancing the performance of a pure 3D network.
	Eventually, it achieves the state-of-the-arts on two large-scale benchmarks (\ie, SemanticKITTI and NuScenes).
	We believe that our work can be applied to a wider range of other scenarios in the future, such as 3D detection and tracking. 
	\\	
	
	{{\noindent\textbf{Acknowledgment.}} 	This work was supported in part by NSFC-Youth 61902335, by the Basic Research Project No. HZQB-KCZYZ-2021067 of Hetao Shenzhen HK S\&T Cooperation Zone, by the National Key R\&D Program of China with grant No.2018YFB1800800, by Shenzhen Outstanding Talents Training Fund, by Guangdong Research Project No. 2017ZT07X152 and No. 2019CX01X104, by the Guangdong Provincial Key Laboratory of Future Networks of Intelligence (Grant No. 2022B1212010001), by the NSFC 61931024\&8192 2046, by NSFC-Youth 62106154, by zelixir biotechnology company Fund, by Tencent Open Fund, and by ITSO at CUHKSZ.}

	\bibliographystyle{splncs}
	\bibliography{0755.bbl}
	\newpage
	
	\setcounter{section}{0}
	\setcounter{figure}{0}
	\setcounter{table}{0}
	\renewcommand\thesection{\Alph{section}}
	
	{\Large\centering \textbf{Supplementary Material}}


	\section{Training and Inference Details}
	\label{implementation}
	%
	For the 3D input, we utilize the widely used data augmentation strategy for semantic segmentation, including global scaling with a random scaling factor sampled from [0.95, 1.05], and global rotation around the Z axis with a random angle.
	For the 2D input, we employ horizontal flipping and color jitter. Each 2D image is cropped to the size 480 $\times$ 320 (width $\times$ height) for faster training.
	The 2DPASS is trained in an end-to-end manner with the SGD optimizer. 
	For the SemanticKITTI validation set, our model was trained with batch size 8 and learning rate 0.24 for 64 epochs, which is kept the same as SPVCNN~\cite{tang2020searching} for fair comparison. 
	For the SemanticKITTI online benchmark, we conduct instance CutMix as \cite{xu2021rpvnet}, and fine-tune the last checkpoint with additional 48 epochs.
	As for the NuScenes dataset, we trained the model with batch size 16 for 80 epochs since the number of points per scene in NuScenes is generally smaller.
	During the inference, following \cite{tang2020searching,zhou2020cylinder3d}, we apply the voting test-time augmentation, \ie, rotating the input scene with 12 angles around the Z axis and averaging the prediction scores.
	All experiments are on Nvidia Tesla V100 GPUs.

	\label{exp}

	\section{Additional Experiments}
	\subsection{Comparing with Multi-Sensor Architecture}
\begin{figure*}[t]
		\begin{centering}
			\includegraphics[width=0.8\textwidth]{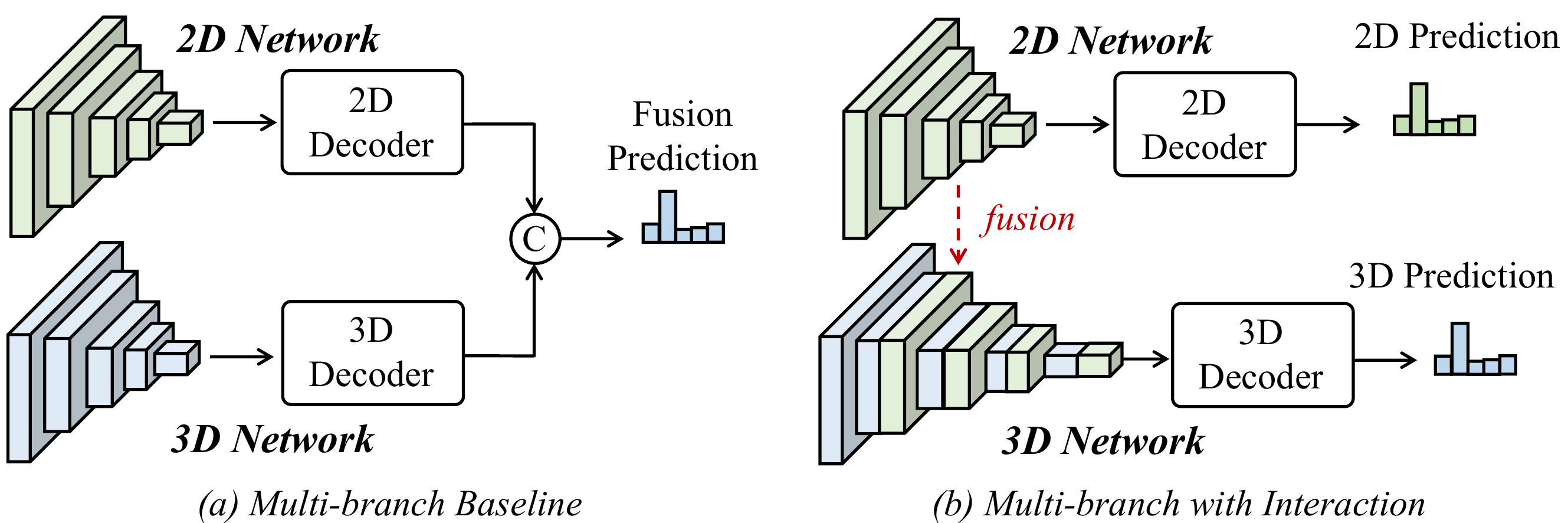}
			
			\caption{{The illustration of multi-sensor methods.\vspace{-.6cm}} 
			}
			\label{fig:supp_figure1}
		\end{centering}	
	
	\end{figure*}

	To further demonstrate the advantages of our 2DPASS upon multi-sensor methods, we set several multi-sensor baselines and compare against them.
	\begin{itemize}
		\item \textbf{PointPainting}: We follow the setup of previous work~\cite{vora2020pointpainting}, which exploits the segmentation logits of images and projects them to the LiDAR space by bird's-eye projection~\cite{yuan2018ocnet} or spherical projection~\cite{milioto2019rangenet++}. 
		Here, we use several pre-trained backbones, \ie,  FCN~\cite{long2015fully} with ResNet34\cite{he2016deep} and DeepLab\_v3~\cite{chen2017rethinking}, to achieve the 2D semantic segmentation logits.
		After that, we use outputs of 2D backbones as the inputs of our 3D network.

		\item \textbf{Multi-branch Baseline}: As shown in Fig.~\ref{fig:supp_figure1}~(a), we design an ensemble architecture through concatenating the output logits from the two modalities.

		\item \textbf{Multi-branch with Interaction}: Instead of only concatenating the predictions, we also concatenate the 2D features from each layer into the corresponding layers in the 3D network, as illustrated in Fig.~\ref{fig:supp_figure1}~(b).
		
		\item \textbf{2DPASS (light)}:  Since above multi-sensor manners are trained with the entire 2D image as input, they are time-consuming and GPU memory cost expensive. So we set all of hidden dimensions as 64 in the 3D network due to GPU memory limitation. This design is different from our manuscript with hidden dimensions 128 due to our light memory cost.
		
	\end{itemize}
	
	The experiment results are shown in Table~\ref{tab:ablation1supp}, where we illustrate the results on NuScenes validation set and inference time (speeds), respectively.
	As shown in Table~\ref{tab:ablation1supp}, using naive combination such as PointPainting~\cite{vora2020pointpainting} and concatenation (\ie, Multi-branch Baseline) of prediction cannot improve the segmentation results obviously while introducing huge computational burden (\ie, there are six $1600\times 900$ camera images corresponding to each point cloud).
	Exploiting feature combination in each scale can slightly improve the performance, but leads to much slower network compared with the pure 3D network.
	On the contrary, 2DPASS (light) achieves the second-best performance in term of mIoU criterion while 60$\times$ speed faster than multi-sensor methods.
	
	\begin{table*}[t]
		\footnotesize
		\vspace{-0.2cm}
		\begin{tabular}{l|cc}
			\hline
			Method & mIoU (\%)  & Speed (ms)\\
			\hline
			PointPainting-FCN-ResNet34 & 76.54  & 2330 \\
			PointPainting-DeepLabV3 & 76.56  & 3347 \\
			Multi-branch Baseline & 77.25  & 2353 \\
			Multi-branch with Interaction & \textbf{79.12}  &  2374 \\\hline
			Basline (light) & 76.04  & 40 \\	
			2DPASS (light) & 78.87  & 40 \\
			\hline
		\end{tabular}%
		\vspace{-0.2cm}
		\caption{Comparison with different multi-sensor manners.}
		\label{tab:ablation1supp}

	\end{table*}

	\begin{table*}[b]
		\small
		\renewcommand\tabcolsep{1.5pt} 
		\caption{Semantic segmentation results on the {NuScenes} valid set.\vspace{-.6cm}}
		
		\begin{center}
			{\fontfamily{cmr}
				\resizebox{\textwidth}{!}{%
					\begin{tabular}{l|c|c|cccccccccccccccc}
						\hline
						
						Method& Input &
						\rotatebox{90}{\bf{mIoU}}&
						\rotatebox{90}{barrier}&
						\rotatebox{90}{bicycle}&
						\rotatebox{90}{bus}&
						\rotatebox{90}{car}&
						\rotatebox{90}{construction~}&
						\rotatebox{90}{motorcycle}&
						\rotatebox{90}{pedestrian}&
						\rotatebox{90}{traffic cone}&
						\rotatebox{90}{trailer}&
						\rotatebox{90}{truck}&
						\rotatebox{90}{driveable}&
						\rotatebox{90}{other flat}&
						\rotatebox{90}{sidewalk}&
						\rotatebox{90}{terrain}&
						\rotatebox{90}{manmade}&
						\rotatebox{90}{vegetation}\\
						\hline
						\hline
						
						\text{(AF)$^2$-S3Net}~\cite{cheng20212} & L &62.2 &60.3  &12.6  &82.3 & 80.0  &  20.1 & 62.0 & 59.0 & 49.0 &  42.2  & 67.4 & 94.2 &  68.0  &  64.1  &  68.6  &  82.9  & 82.4  \\
						\text {AMVNet}~\cite{liong2020amvnet} & L & 76.1 & \bf{79.8} & 32.4 & 82.2 & 86.4 & \bf{62.5} & 81.9 & 75.3 & \bf{72.3} & \bf{83.5} & 65.1 & \bf{97.4} & 67.0 & \bf{78.8} & 74.6 & 90.8 & 87.9 \\
						\text {Cylinder3D}~\cite{zhou2020cylinder3d} & L & 76.1 & 76.4 & 40.3 & 91.2 & \bf{93.8} & 51.3 & 78.0 & 78.9 & 64.9 & 62.1 & 84.4 & 96.8 & 71.6 & 76.4 & 75.4 & 90.5 & 87.4 \\
						\text{RPVNet}~\cite{xu2021rpvnet} & L & 77.6 & 78.2 & 43.4 & 92.7 & 93.2 & 49.0 & \bf{85.7} & 80.5 & 66.0 & 66.9 & 84.0 & 96.9 & 73.5 & 75.9 & 76.0 & 90.6 & 88.9 \\
						\text{PMF}~\cite{zhuang2021perception} & L+C & 76.9 & 74.1 & 46.6 & 89.8 & 92.1 & 57.0 & 77.7 & 80.9 & 70.9 & 64.6 & 82.9 & 95.5 & 73.3 & 73.6 & 74.8 & 89.4 & 87.7 \\
						\text{2D3DNet}~\cite{genova2021learning} & L+C & 79.0 & 78.3 & \bf{55.1} & 95.4  &  87.7  &  59.4  &  79.3  &  80.7  &  70.2  &  68.2  &  86.6  &  96.1  &  \bf{74.9}  &  75.7  &  75.1  & \bf{91.4}  &  \bf{89.9} \\\hline
						\text{Baseline} & L & 76.2 & 75.3  & 43.5  & 95.3  & 91.2  & 54.5  & 78.9  & 72.8  & 62.1  & 70.0  & 83.2  & 96.3  & 73.2  & 74.2  & 74.9  & 88.1  & 85.9\\
						\text{\bf{2DPASS(Ours)}}  & L& \bf{79.4} & 78.8 & 49.6 & \bf{95.6} & 93.6 & 60.0 & 84.1 & \bf{82.2} & 67.5 & 72.6 &\bf{88.1} & 96.8 & 72.8 & 76.2 & \bf{76.5} & 89.4 & 87.2\\
						\hline
			\end{tabular} }}
		\end{center}
		\label{tab:nus_seg_valid}
	\end{table*}

%

%
%

	\subsection{Concrete Results}
	\label{results}
	In this section, we give our detailed results on the NuScenes dataset in Table~\ref{tab:nus_seg_valid} as a benchmark for future work.
%
%
%
%
%
%
	
\end{document}